\documentclass[sigconf,review=false,anonymous=false]{acmart}
\renewcommand\footnotetextcopyrightpermission[1]{}
\acmConference{}{}{}

\usepackage{subcaption}

\AtBeginDocument{%
  }

\newcommand{\sobol}{Sobol\kern-0.15em\'{ } }

\begin{document}

\title{Class Incremental Learning for Algorithm Selection}

\author{Mate Botond Nemeth}
%\authornotemark[2]
\affiliation{%
  \institution{Edinburgh Napier University}
  \city{Edinburgh}
  \country{Scotland, UK}}
  \email{mate.nemeth@napier.ac.uk}

\author{Emma Hart}
\affiliation{%
  \institution{Edinburgh Napier University}
  \city{Edinburgh}
  \country{Scotland, UK}}
  \email{e.hart@napier.ac.uk}

\author{Kevin Sim}
\affiliation{%
\institution{Edinburgh Napier University}
  \city{Edinburgh}
  \country{Scotland, UK}}
  \email{k.sim@napier.ac.uk}
  
\author{Quentin Renau}
\affiliation{%
\institution{Edinburgh Napier University}
  \city{Edinburgh}
  \country{Scotland, UK}}
  \email{q.renau@napier.ac.uk}

\renewcommand{\shortauthors}{ Nemeth et al.}

\begin{abstract}
Algorithm selection is commonly used  to predict the best solver from a portfolio per per-instance. In many real scenarios, instances arrive in a stream:  new instances become available over time, while the number of class labels can also grow as new data distributions arrive downstream. As a result, the classification model needs to be periodically updated to reflect additional solvers without catastrophic forgetting of past data.  In machine-learning (ML), this is referred to as \textit{Class Incremental Learning (CIL)}. While commonly addressed in ML settings, its relevance to algorithm-selection in optimisation has not been previously studied. Using a bin-packing dataset, % from combinatorial optimisation domains, 
we benchmark $8$ continual learning methods with respect to their ability to
withstand catastrophic forgetting. We find that \textit{rehearsal-based} methods
%that save exemplars of previous data and use them when retraining 
significantly outperform other CIL methods. While there is evidence of forgetting, the loss is small at around $7\%$. %Overall accuracy of the final model at the end of the stream is $\ge 82\%$. 
Hence, these methods appear to be a viable approach to continual learning in streaming optimisation scenarios.

\end{abstract}


\maketitle

\section{Introduction}
In many optimisation settings, the need to perform  per-instance algorithm selection to choose the best solver per instance is well known ~\cite{kerschke2019automated}. 
However, the vast majority of these approaches that employ some form of classifier as a selector make two assumptions: (1) there is a fixed portfolio of solvers (and therefore a fixed set of labels to predict); (2) available training data contains  some samples from all classes.

However, in the real-world,  these assumptions often do not hold. Instances often arrive continually in a stream \cite{gomes2017survey}. Kerschke {\em et. al.}~\cite{kerschke2019automated} note that `the importance of automated algorithm selection and related approaches in the context of learning on streaming data should not be neglected', but it is acknowledged that this is challenging for current methods ~\cite{bifet2018machine}. For example,
when a model is initially trained, the available data from the stream may only contain examples representing a subset of class labels. Furthermore, in typical scenarios, data points can only be evaluated once and are discarded afterwards~\cite{bifet2018machine}, while in many cases, instance data cannot be held indefinitely, either due to storage constraints \cite{krempl2014open} or to  privacy issues \cite{de2021continual}.  If a classification model needs to be updated due to new instances arriving  with new class labels,  previously seen data may no longer be accessible for inclusion in training. This introduces a signifcant challenge: how to update a model to include new classes without erasing knowledge of previous classes. This is known as \textit{catastrophic forgetting}. Within the machine-learning community, the field of \textit{Continual Learning} arose in response to this challenge~\cite{parisi2019continual,van2022three}. In its most general sense, this is defined as \textit{"an adaptive algorithm capable of learning from a continuous stream of information, with such information becoming progressively available over time and where the number of tasks to be learned are not predefined"}~\cite{parisi2019continual}. 
Here, we specifically focus on \textit{Class Incremental Learning (CIL)}, that is, the case where a classifier must incrementally learn to discriminate between a growing number of classes~\cite{zhou2024class}.

%%%
With the advent of Deep Learning, many  approaches to CIL that address the issue of catastrophic forgetting have been proposed~\cite{zhou2024class}. While there are many examples of practical applications in the context of ML,% particularly in the field of image recognition~\cite{Zhou2023Class-IncrementalSurvey},
to the best of our knowledge, \textit{there has been no attempt to evaluate how these methods perform when applied in updating an algorithm-selector over time in the context of streaming optimisation problems where new classes appear over time}. We therefore perform a benchmark analysis of eight state-of-the-art CIL techniques on data from online bin-packing %, an example of a combinatorial optimisation problem, 
, where instances are solved with a heuristic selected from a portfolio of deterministic heuristic solvers.

\section{Background: Class Incremental Learning}
In a class-incremental learning scenario, an algorithm must incrementally learn to distinguish between a growing number of objects or classes~\cite{van2022three}.
For example, consider an algorithm selection scenario in which a model first learns when solvers $S1,S2$ are useful; later it learns when $S3, S4$ are useful but also must be able to  distiniguish between (say) $S1$ and $S4$. The latter scenario of learning to discriminate between classes that are not observed together is particularly challenging for deep neural networks, especially if storing examples of previously seen classes is not allowed~\cite{zhou2024class}.

\subsection{Learning strategies}
There are multiple different approaches for CIL~\cite{zhou2024class}. In this work 8 popular methods were chosen from the state-of-the-art, representing examples from four  different categories  of methods, according to the taxonomoy outlined in~\cite{zhou2024class}.
\textit{Parameter regularisation} methods aim to estimate the importance of each parameter in a model and when training on new data, try to penalise changes being made to them. 
We have used \textit{Elastic Weight Consolidation } (EWC)~\cite{Kirkpatrick2016OvercomingNetworks}, \textit{Memory Aware Synapsis} (MAS)~\cite{Aljundi2017MemoryForget}, \textit{Synaptic Intelligence }(SI)~\cite{Zenke2017ContinualIntelligence}. 
\textit{ Data-regularization} methods  use exemplars from previous tasks as indicators to regularize model updating:
\textit{Gradient Episodic Memory} (GEM)~\cite{Lopez-Paz2017GradientLearning}, \textit{Average Gradient Episodic Memory }(AGEM)~\cite{Chaudhry2018EfficientA-GEM}.
While \textit{Data-replay} methods revisit exemplars from previous tasks to mitigate catastrophic forgetting:
\textit{Feature Replay} (FR)~\cite{magistri2024elastic}, \textit{Experience Replay}~\cite{ratcliff1990connectionist}.
\textit{Knowledge distillation} approaches enable knowledge transfer from the last trained model to the new model while training on the new task from this category we have used
\textit{Learning without Forgetting} (LwF)~\cite{Li2016LearningForgetting}.

The reader is referred to the supplementary information \cite{dataCIL} and ~\cite{zhou2024class} for full details. 
\section{Methodology}

% explain task = set of labels
Formally, we define the problem addressed as follows, using  the standard nomenclature of CIL. We
assume there is a sequence of $\mathcal{B}$ training tasks ${\mathcal{D}_1,\mathcal{D}_2,...,\mathcal{D}_\mathcal{B}}$, where each task consists of a dataset of $n_b$ training instances associated with $\mathcal{Y}_b$ labels. 
In the CIL scenario studied,  $\mathcal{Y}_b \cap \mathcal{Y}_{b'} = \varnothing$ for $b \neq b'$, i.e., labels are different for each training tasks. 
In all experiments, we set $\mathcal{B}=2$ and $\mathcal{Y}_b=2$, i.e. 2 labels are observed for each dataset $\mathcal{D}_\mathcal{B}$.

\subsection{Datasets and Solver Portfolios}

We use online-binpacking to demonstrate CIL in an algorithm-selection setting. 
The dataset consists of $4000$ instances first published in ~\cite{alissa2019algorithm}. Each instance consists of $120$ items to be packed, with items drawn from a uniform distribution in the range $[20,100]$. 
The dataset is associated with $4$ deterministic heuristic solvers, each of which `wins' exactly $1000$ of the instances. Here, a `win' denotes that a heuristic obtains the best packing according to the Falkanauer metric: this metric is defined in~\cite{220088} and prefers `well-filled' bins over a set of equally filled bins. 
The Falkanauer metric is bounded between $[0,1]$ and aims to be maximised. Four deterministic solvers are considered: \textbf{First-Fit} (FF), \textbf{Best-Fit }(BF), \textbf{Next-Fit} (NF), \textbf{Worst-Fit} (WF).

\subsection{Evaluation Protocol}
The architecture and the hyperparameters of the classifier model are in the supplementary information \cite{dataCIL}.
All CIL learning methods are implemented using the Avalanche library\cite{Carta2023Avalanche:Learning}. 

The model is first trained on task $\mathcal{D}_1$ which has 2 class labels and train it using one of the CIL methods.  The same CIL method is then applied to update the model using data from $\mathcal{D}_2$.  This data also has 2 class labels (distinct from $\mathcal{D}_1$).  When training on $\mathcal{D}_2$, the model does not have access to the training data from $\mathcal{D}_1$, except in rehearsal-based strategies where a select number of exemplars are saved. We set this number to 100 as this corresponds to the minimum number of instances associated with a class. This exemplar set is usually considerably smaller than the size of the $\mathcal{D}1$ dataset.

There are $4$ possible class labels in total when considering the entire stream: we conduct experiments for each possible construction of $\mathcal{D}_1, \mathcal{D}_2$, resulting in $6$ different streams, each with two tasks both associated with two solvers.
%where 2 solvers are selected for the first experience and the remaining 2 for the second
We evaluate the influence of the size of the training set on model accuracy. Each stream consisting of $\mathcal{D}_1,\mathcal{D}_2$  using 5 different sizes of training data, where $n \in \{100, 200, 300, 400, 500\}$ instances per solver. 
The test data included all of the remaining available instances. 
Results are aggregated over $30$ runs (6 combinations of  $\mathcal{D}_1, \mathcal{D}_2$, $5$ training sizes). The total number of training epochs per dataset is identical in all training methods. 
The parameters for each CIL methods can be seen in the supplementary information \cite{dataCIL}.

\subsubsection{Metrics and Comparisons}

We record the following metrics over the $30$ runs described above: accuracy of classifier on task $\mathcal{D}_1$; accuracy of classifier on task $\mathcal{D}_1$ after retraining on $\mathcal{D}_2$; accuracy of classifier on task $\mathcal{D}_2$ after retraining on $\mathcal{D}_2$; accuracy of retrained classifier on all 4 classes.

We also calculate two `hypothetical'  metrics to place the results in context. For this, we assume a\textit{ non-streaming} scenario in which \textit{all} data ever observed in the stream is available immediately to train a selector. We then calculate two metrics based on the accuracy of two selectors: \textit{Oracle-Selector}: a classifer trained on examples of data from  both $\mathcal{D}_1$ and $\mathcal{D}_2$; \textit{Cumulative-Selector}: a classifier is first trained $\mathcal{D}_1$ only, and then training continued with an expanded training set consisting of \textit{all} examples from $\mathcal{D}_1$ \textit{and} $\mathcal{D}_2$.

Note that the oracle-selector is equivalent to a classical selector when data for all classes is available for training and is not applicable in a streaming scenario.
Nevertheless, it provides some context for our results in terms of what accuracy might be achieved if all data had been available. The cumulative selector is effectively an extreme version of CIL in which \textit{all} training data from $\mathcal{D}_1$ is available when training $\mathcal{D}_{2}$. Again, this is an unrealistic benchmark but provides context for the results.

\begin{table*}[]
\caption{Bin-Packing CIL mean accuracies with standard deviation.}
\vspace{-1.8mm}
\label{tab:bp_all}
\begin{tabular}{cllll}
\toprule
\textbf{Method}     & \multicolumn{1}{c}{\textbf{\begin{tabular}[c]{@{}c@{}}Training on $\mathcal{D}1$; \\ $\mathcal{D}1$ Accuracy\end{tabular}}} & \multicolumn{1}{c}{\textbf{\begin{tabular}[c]{@{}c@{}}Training on $\mathcal{D}2$ ; \\ $\mathcal{D}1$  Accuracy\end{tabular}}} & \multicolumn{1}{c}{\textbf{\begin{tabular}[c]{@{}c@{}}Training on $\mathcal{D}2$ ; \\ $\mathcal{D}2$  Accuracy\end{tabular}}} & \multicolumn{1}{c}{\textbf{$\mathcal{D}1$, $\mathcal{D}2$ Accuracy}} \\
\midrule
\textbf{Oracle}     & N/A                                                      & N/A                                                      & N/A                                                      & 0.883 (0.032)                                  \\
\textbf{Cumulative} & 0.907 (0.064)                                            & 0.834 (0.117)                                            & 0.865 (0.071)                                            & 0.848 (0.054)                                  \\
\midrule
\textbf{EWC}        & 0.93 (0.041)                                             & 0.006 (0.027)                                            & 0.909 (0.088)                                            & 0.458 (0.048)                                  \\
\textbf{MAS}        & 0.913 (0.087)                                            & 0.036 (0.087)                                            & 0.898 (0.068)                                            & 0.467 (0.062)                                  \\
\textbf{SI}         & 0.924 (0.054)                                            & 0.007 (0.036)                                            & 0.922 (0.064)                                            & 0.464 (0.037)                                  \\
\textbf{LwF}        & 0.924 (0.057)                                            & 0.005 (0.027)                                            & 0.91 (0.068)                                             & 0.458 (0.037)                                  \\
\textbf{GEM}        & 0.912 (0.052)                                            & 0.683 (0.084)                                            & 0.916 (0.067)                                            & 0.798 (0.034)                                  \\
\textbf{AGEM}       & \textbf{0.933 (0.053)}                                            & 0.442 (0.236)                                            & 0.938 (0.041)                                            & 0.69 (0.117)                                   \\
\textbf{FR}         & 0.928 (0.048)                                            & 0.043 (0.1)                                              & \textbf{0.955 (0.037)}                                            & 0.499 (0.048)                                  \\
\textbf{Replay}     & 0.928 (0.045)                                            & \textbf{0.739 (0.084)}                                           & 0.906 (0.088)                                            & \textbf{0.823 (0.036)}                                 \\
\bottomrule
\end{tabular}
\end{table*}

\begin{table}[]
\caption{Accuracy per class in $\mathcal{D}1$ trained using Replay on $\mathcal{D}1$ followed by training on $\mathcal{D}2$.}
\vspace{-1.8mm}
\label{tab:D1_acc_loss}
\begin{tabular}{llll}
\toprule
\multicolumn{1}{c}{\textbf{$\mathcal{D}1$}} & \multicolumn{1}{c}{\textbf{Model($\mathcal{D}1$)}} & \multicolumn{1}{c}{\textbf{Model($\mathcal{D}2$)}} & \multicolumn{1}{c}{\textbf{Loss per $\mathcal{D}1$ class}} \\
\midrule
{[}BF, FF{]}                            & {[}0.86, 0.94{]}                            & {[}0.91, 0.74{]}                            & {[}0.05, -0.19{]}                           \\
{[}BF, NF{]}                            & {[}0.89, 0.92{]}                            & {[}0.9, 0.94{]}                             & {[}0.01, 0.01{]}                            \\
{[}BF, WF{]}                            & {[}0.91, 0.93{]}                            & {[}0.77, 0.83{]}                            & {[}-0.14, -0.1{]}                           \\
{[}FF, NF{]}                            & {[}0.9, 0.95{]}                             & {[}0.75, 0.94{]}                            & {[}-0.15, -0.01{]}                          \\
{[}FF, WF{]}                            & {[}0.96, 0.94{]}                            & {[}0.61, 0.75{]}                            & {[}-0.35, -0.18{]}                          \\
{[}NF, WF{]}                            & {[}0.91, 0.96{]}                            & {[}0.9, 0.78{]}                             & {[}-0.01, -0.17{]}                          \\

\bottomrule
\end{tabular}
\end{table}

\begin{table}[]
\caption{Difference in accuracy with the final model using Replay for each class in a pair, depending on whether the pair is $\mathcal{D}1$ training data or $\mathcal{D}2$. Difference is expressed as $\mathcal{D}1$ accuracy minus $\mathcal{D}2$ accuracy for each class.}
\vspace{-1.8mm}
\label{tab:D1_D2_classes}
\begin{tabular}{llll}
\toprule
\multicolumn{1}{c}{\textbf{Classes}} & \multicolumn{1}{c}{\textbf{\begin{tabular}[c]{@{}c@{}}Classes used in\\ training as $\mathcal{D}1$\end{tabular}}} & \multicolumn{1}{c}{\textbf{\begin{tabular}[c]{@{}c@{}}Classes used in\\ training as $\mathcal{D}2$\end{tabular}}} & \multicolumn{1}{c}{\textbf{Difference}} \\
\midrule
{[}BF, FF{]}                         & {[}0.91, 0.74{]}                & {[}0.92, 0.76{]}                & {[}-0.01, -0.02{]}                      \\
{[}BF, NF{]}                         & {[}0.9, 0.94{]}                 & {[}0.91, 0.91{]}                & {[}-0.01, 0.02{]}                       \\
{[}BF, WF{]}                         & {[}0.77, 0.83{]}                & {[}0.89, 0.68{]}                & {[}-0.12, 0.14{]}                       \\
{[}FF, NF{]}                         & {[}0.75, 0.94{]}                & {[}0.73, 0.94{]}                & {[}0.02, 0.0{]}                         \\
{[}FF, WF{]}                         & {[}0.61, 0.75{]}                & {[}0.72, 0.69{]}                & {[}-0.1, 0.06{]}                        \\
{[}NF, WF{]}                         & {[}0.9, 0.78{]}                 & {[}0.92, 0.75{]}                & {[}-0.02, 0.03{]}                       \\
\bottomrule
\end{tabular}
\end{table}

\begin{figure}
  \centering

  \includegraphics[width=\linewidth]{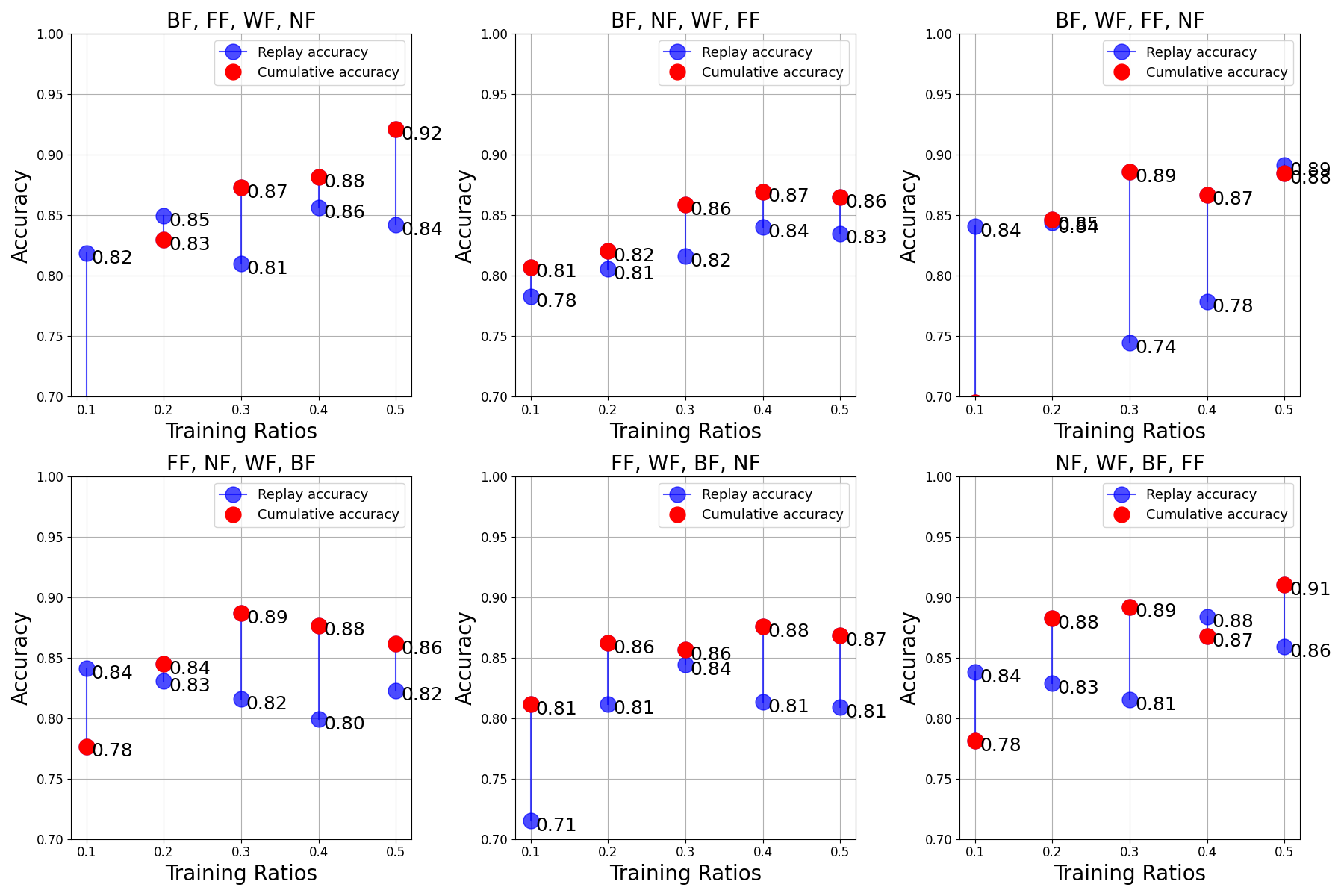}
  %\Description{Cumulative and Replay accuracy comparison for the binapcking model}
  \label{fig:bp_replay_cumulative_comp}
\hfill
\vspace{-6mm}
\caption{Comparison of Cumulative and Replay accuracy accross all $\mathcal{D}_1, \mathcal{D}_2$ combinations and training sizes. Red dots - cumulative model; blue dots - Replay model; line length indicates the magnitdue of the loss. 
}
\label{fig:allCumulative}
\end{figure}

\begin{figure}
  \centering

  \includegraphics[width=\linewidth]{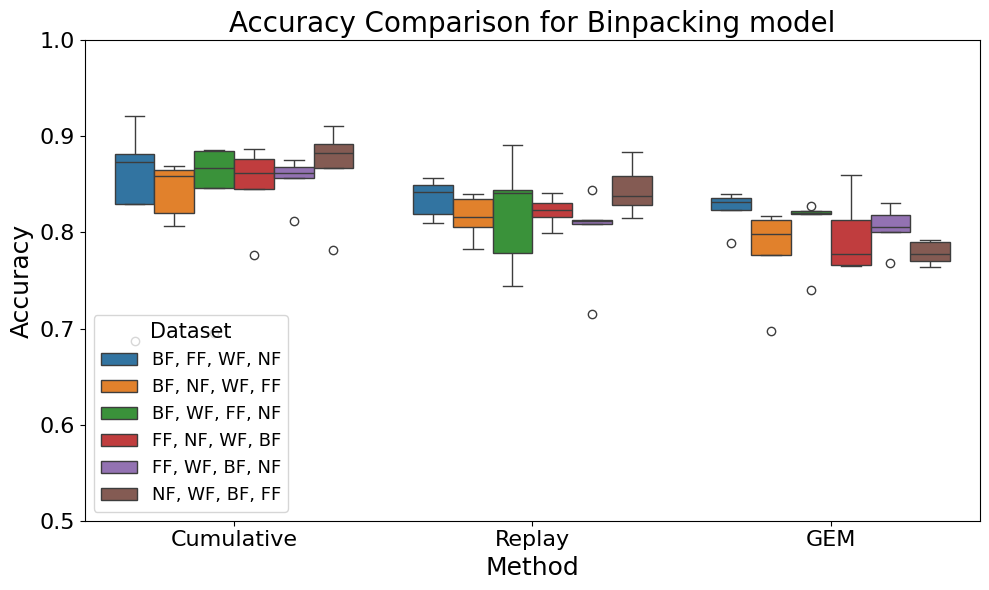}
  \Description{Plot of all $\mathcal{D}_1, \mathcal{D}_1$ combination accuracies for bin-packing}
  \label{fig:bp_cumulative_boxplot}

 \centering
\vspace{-6mm}
\caption{Accuracy over all $4$ classes separated by the order in which pairs of classes are presented. Results are shown for the cumulative method as context, and the best performing CIL methods Replay and GEM. %Order of presentation shown in the key.
}
\label{fig:boxplotsPerOrder}
\end{figure}

\section{Results}

We first present results that evaluate the robustness of the different CIL methods to catastrophic forgetting.  These results are aggregated over all combinations of training set size and order of presentation of tasks and provide an overall summary. We then consider the impact of the size of the training data and specific order in which tasks in the stream  are presented, analysing the impact of these factors separately.

\subsection{Comparison of CIL methods}
Table ~\ref{tab:bp_all} shows the mean accuracy and standard deviation after training on $\mathcal{D}1$ and training on $\mathcal{D}2$.

All training strategies achieved similar results when training on $\mathcal{D}1$, since they train on the first observed classes in the same way.
After training on $\mathcal{D}_2$ most methods undergo catastrophic forgetting w.r.t $\mathcal{D}_1$. The Replay method results in the smallest loss of  $18.9\%$ when compared to $\mathcal{D}_2$ and GEM follows closely with $22.9\%$. On the other hand, the Cumulative model that assumes knowledge of all data exhibits a loss of only $7\%$ on the $\mathcal{D}_1$ classes after retraining on $\mathcal{D}_2$. We note that after training on $\mathcal{D}_2$,  all of the CIL methods result in better accuracy with respect to $\mathcal{D}_2$ than the  Cumulative model (a $9\%$  difference in accuracy between the best CIL model FR and the reference model). Finally, considering the accuracy of the retrained model with respect to seen classes we find the best result from Replay method that has an $82.3\%$ over all classes. This result is very close to the result obtained from the Cumulative model ($84\%$), despite the fact that the replay methods only uses $100$ exemplars in the retraining phase from $\mathcal{D}1$. The Oracle training performs better ($88\%$) than both the Cumulative and Replay method; however, this result is unattainable in a CIL setting. The Parameter Regularisation and Knowledge Distillation methods are significantly worse. Feature Replay, despite being a Data-Replay method, achieves similar accuracies as Parameter Regularisation and Knowledge distillation. The second Data-Regularisation method AGEM contrary to GEM also suffers from catastrophic forgetting.

\subsection{Impact of Training set size and order of task presentation}

\paragraph{Comparison of replay method to the cumulative training} 
In Figure~\ref{fig:allCumulative} we visualise the loss in accuracy from the best CIL method Replay from the final trained model ($\mathcal{D}_1$, $\mathcal{D}2$) in comparison to the hypothetical Cumulative training case (recall this makes all training data from both tasks available). The  figures are separated according to the order in which tasks are presented; in each figure, we also show the difference in accuracy obtained based on the size of the training set. 
% \quentin{maybe add it to the caption instead ?}
Cumulative training struggles on some $\mathcal{D}_1$, $\mathcal{D}2$ combinations for small training sizes (see for example the combination where $\mathcal{D}_1$ is (BF,FF) and the combination where $\mathcal{D}_1$ is (FF,NF)).
In general there are no obvious trends with respect to any correlation between terms of the size of the training set and loss in accuracy across the different presentations of the classes.

\paragraph{Extent of forgetting}

In Table~\ref{tab:D1_acc_loss} we show the accuracy \textit{per class} for $\mathcal{D}1$  after training on $\mathcal{D}1$ and then after retraining with $\mathcal{D}2$, for each of the $6$ possible  versions of the  $\mathcal{D}1$ dataset. Using the Replay method, we observe that $4$ of the $\mathcal{D}1$ datasets have reduced accuracy on both classes after retraining. Interestingly, when the $\mathcal{D}1$ dataset is (BF,NF), the \textit{retrained} model has slightly improved accuracy on both of these classes compared to the original model.

\subsection{Impact of ordering of data in a stream}

As data in a stream varies over time, it is possible that the order in which data appears can affect performance. To investigate this, we analyse performance on each of the $6$ streams created.  
Figure~\ref{fig:boxplotsPerOrder} shows that the different combinations of $\mathcal{D}1$ and $\mathcal{D}2$ can result in varying performances (each box is plotted over $5$ different training ratios). This applies to both the cumulative setting and the two best CIL strategies (Replay, GEM) shown.
%The variation between combinations indicates a variation in the datasets themselves and in the relative ability of the models to distinguish certain pairs of solvers.
We provide further insight into the order in which classes are presented by  considering whether a dataset is better classified when presented as $\mathcal{D}_1$ or $\mathcal{D}_2$. Table~\ref{tab:D1_D2_classes} shows the accuracy of the two named classes according to the \textit{final model} (i.e. with knowledge of all $4$ classes). Due to the tendency for retrained models to forget past data, we would expect that the accuracy of a pair of classes when presented as $\mathcal{D}2$ would be better than when presented as $\mathcal{D}1$. However, for five streams, we see a reduction in accuracy in \textit{one} of the classes when the pair is presented as $\mathcal{D}2$, usually at the expense of an increase in accuracy in the second class of the pair. 
The difference is most pronounced for the (BF,WF) pair, in which the accuracy of the WF class reduces by $14\%$ when it is presented as $\mathcal{D}2$. 

\section{Conclusion}

We find that the class of CIL methods known as \textit{rehearsal-based} methods that save exemplars or prototypes of previous data and use them in retraining perform significantly better than other categories of CIL when evaluated over multiple streams. While the model suffers from some forgetting, the accuracy at the end of stream with respect to 4 solver classses is $82.3\%$. The results compare favourably to those obtained in a \textit{non-streaming }scenario when all data and class labels are known from the start, where the loss compared to this benchmark is $6\%$. The size of the training set used to (re)train each model does not significantly impact results. In contrast, the order in which classes appear in a stream can impact results, suggesting that some pairs of classes are more difficult to distinguish than others.

All training data is available at~\cite{dataCIL}.

% \begin{acks}
% Author XXX is funded by YYYYY
% \end{acks}

% \vspace{-1.5mm}
% \newpage
\bibliographystyle{ACM-Reference-Format}
\bibliography{extraRefs}

% \appendix

\end{document}